%% file: main.tex
\documentclass[10pt,twocolumn,letterpaper]{article}

\usepackage{cvpr}              

\usepackage{algorithm}
\usepackage{algorithmic}
\usepackage{amsmath}
\usepackage{amssymb} 
\usepackage{booktabs}
\usepackage{multirow}
\usepackage{xcolor}
\usepackage{graphicx}

\input{preamble}
\definecolor{cvprblue}{rgb}{0.21,0.49,0.74}
\usepackage[pagebackref,breaklinks,colorlinks,allcolors=cvprblue]{hyperref}

\def\paperID{14045} 
\def\confName{CVPR}
\def\confYear{2026}

\title{EMAD: Evidence-Centric Grounded Multimodal Diagnosis for Alzheimer’s Disease}

\author{Qiuhui Chen$^{1}$, Xuancheng Yao$^{2}$, Zhenglei Zhou$^{3}$, Xinyue Hu$^{2}$, Yi Hong$^{2}$\thanks{Corresponding author.}\\
$^{1}$East China University of Science and Technology, $^{1}$Shanghai Jiao Tong University, $^{3}$Tencent\\
{\tt\small chenqh@ecust.edu.cn, \{2212582443, hxy1246475237, yi.hong\}@sjtu.edu.cn, rianzhou@tencent.com}
}

\begin{document}
\maketitle
\input{sec/0_abstract}    
\input{sec/1_intro}
\input{sec/2_rela}

\input{sec/3_method}

\input{sec/4_exp}

\input{sec/5_conclusion}

{
    \small
    \bibliographystyle{ieeenat_fullname}
    \bibliography{main}
}


\end{document}

%% file: sec/0_abstract.tex
\begin{abstract}

Deep learning models for medical image analysis often act as “black boxes,” seldom aligning with clinical guidelines or explicitly linking decisions to supporting evidence. This is especially critical in Alzheimer's disease (AD), where predictions should be grounded in both anatomical and clinical findings. We present \textbf{EMAD}, a vision–language framework that generates structured AD diagnostic reports with each claim explicitly grounded in multimodal evidence. EMAD uses a hierarchical \textbf{Sentence–Evidence–Anatomy (SEA) Grounding} mechanism: (i) sentence-to-evidence grounding links generated sentences to clinical evidence phrases, and (ii) evidence-to-anatomy grounding localizes corresponding structures on 3D brain MRI. To reduce dense annotation requirements, we propose \textbf{GTX-Distill}, which transfers grounding behavior from a teacher trained with limited supervision to a student operating on model-generated reports. We further introduce \textbf{Executable-Rule GRPO}, a reinforcement fine-tuning scheme with verifiable rewards that enforces clinical consistency, protocol adherence, and reasoning–diagnosis coherence. On the AD-MultiSense dataset, EMAD achieves state-of-the-art diagnostic accuracy and produces more transparent, anatomically faithful reports than existing methods; we will release code and grounding annotations to support future research in trustworthy medical vision–language models.

\end{abstract}

%% file: sec/1_intro.tex
\section{Introduction}

Alzheimer's disease (AD) is a heterogeneous syndrome involving structural brain changes, cognitive decline, genetic risk, and fluid biomarkers. In clinical practice, physicians integrate structural MRI (sMRI), neuropsychological tests, APOE genotype, cerebrospinal fluid (CSF) markers, demographics, and comorbidities for diagnosis~\cite{frisoni2010clinical,lautner2014apolipoprotein}. In contrast, many AI-based approaches still operate on isolated modalities~\cite{jang2022m3t,ohman2021current}, missing cross-modal dependencies and risking incomplete or biased conclusions. This motivates multimodal models that jointly reason over imaging and clinical data for more robust AD characterization~\cite{venugopalan2021multimodal,chen2024alifuse,chen2024smart,yang2024alzheimer}.

\begin{figure}[tbp]
    \centering
    \includegraphics[width=\linewidth]{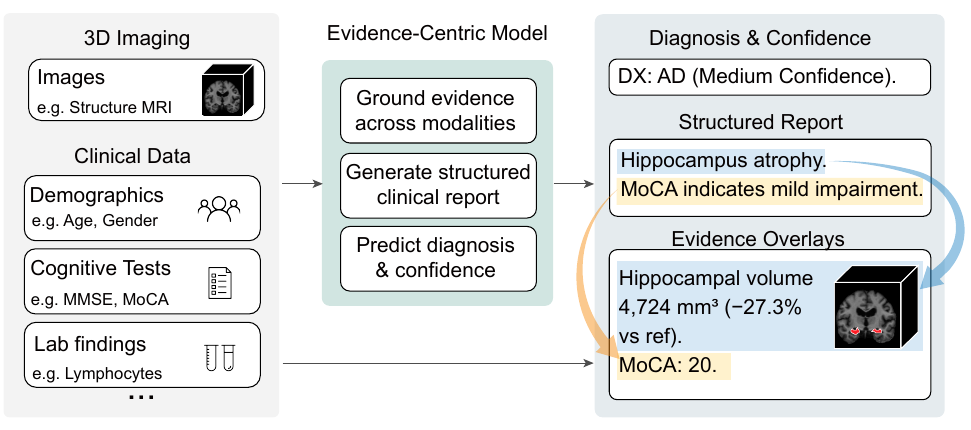}
    \caption{Evidence-centric multimodal clinical diagnosis.}
    \label{fig:abstract}
\end{figure}

Recent multimodal models for AD integrate sMRI, cognitive scores, genetics, and biomarkers~\cite{chen2024alifuse,zhou2023transformer}, but typically behave as black boxes, outputting labels or risk scores without transparent reasoning or explicit links to supporting evidence. This is problematic in settings such as differentiating AD from overlapping neurodegenerative conditions, where clinicians must understand \emph{why} a decision was made and which findings support it. Multimodal large language models (MLLMs)~\cite{openai2023gpt4,grattafiori2024llama} provide a natural interface for report-like outputs, yet current medical uses~\cite{bai2024m3d} rarely (i) link generated sentences to specific clinical entries, (ii) localize claims to 3D brain anatomy, or (iii) enforce adherence to diagnostic frameworks such as NIA-AA.

We introduce EMAD, an end-to-end vision--language framework for \textit{transparent, evidence-grounded, and anatomically faithful} AD reporting. Given a patient's 3D sMRI and structured or semi-structured clinical variables (demographics, genetics, cognitive tests, CSF biomarkers, and other assessments), EMAD generates a structured diagnostic report and a calibrated diagnosis (Fig.~\ref{fig:abstract}). Multimodal encoders and a bidirectional cross-attention fusion module align neuroimaging and clinical representations, followed by a causal language model that produces the report. A Sentence--Evidence--Anatomy (SEA) Grounding head enforces a hierarchical chain of accountability: each sentence is grounded to explicit clinical evidence, which is further grounded to localized anatomy in the 3D MRI via evidence-conditioned segmentation.

Directly annotating sentence--evidence links and anatomy masks is expensive, so we propose GTX-Distill, a label-efficient grounding transfer strategy. A teacher grounder is trained on a small subset with dense grounding supervision, and EMAD distills its soft distributions over evidence and anatomy into a student grounder using model-generated reports via KL-based distillation. To further improve clinical faithfulness, we perform reinforcement fine-tuning with Executable-Rule Group Relative Policy Optimization (GRPO), an RL-with-verifiable-rewards (RLVR) scheme that uses executable rewards to enforce structured reporting, NIA-AA–consistent diagnostic categories and biomarker thresholds, and reasoning--diagnosis consistency, while a KL regularizer anchors the policy to a supervised reference model.
Our main contributions are:
\begin{itemize}
    \item EMAD, an end-to-end multimodal vision--language framework that generates structured AD diagnostic reports with calibrated diagnoses, explicitly grounded to clinical evidence fields and localized 3D anatomy.
    \item SEA Grounding with GTX-Distill, a hierarchical sentence$\rightarrow$clinical$\rightarrow$anatomy grounding module and label-efficient distillation scheme that transfers grounding behavior from a teacher trained on limited annotations to a student operating on large-scale generated reports.
    \item Executable-Rule GRPO, an RLVR-based fine-tuning stage with executable rewards that enforce structured output, NIA-AA–consistent decisions, and reasoning--diagnosis entailment, enabling clinically faithful, interpretable AD diagnosis across large-scale cohorts.
\end{itemize}

%% file: sec/2_rela.tex
\begin{figure*}[t]
    \centering
    \includegraphics[width=\linewidth]{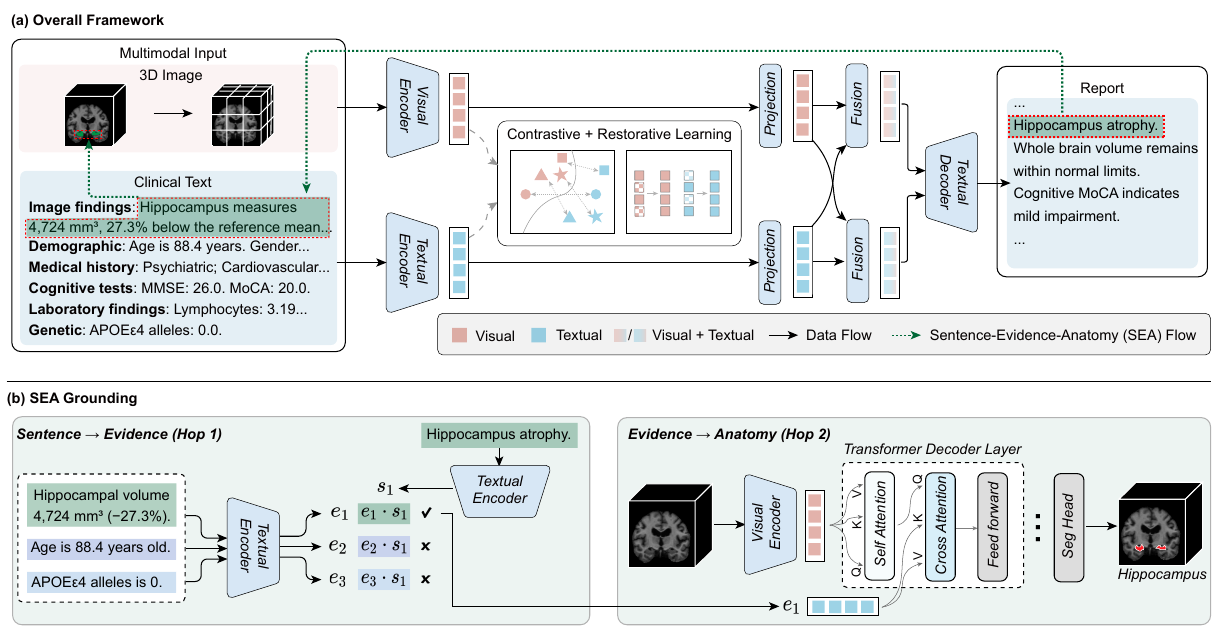}
    \caption{
    (a) Overall framework from multimodal inputs (3D image, clinical text) to encoders, linear projections, multimodal fusion, and textual decoding to a report.
    (b) SEA Grounding produces evidence and anatomy distributions via two-step hierarchical alignment.
    }
    \label{fig:overview}
\end{figure*}

\begin{figure}[htbp]
    \centering
    \includegraphics[width=0.9\linewidth]{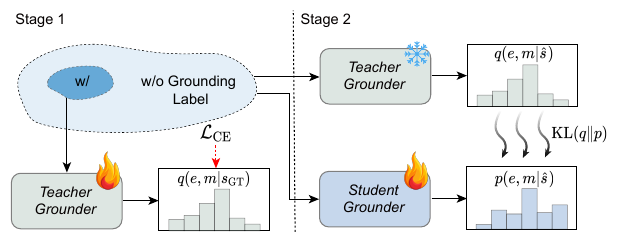}
    \caption{Grounding Transfer Distillation (GTX Distill).}
    \label{fig:gtx}
\end{figure}

\section{Related Works}

\noindent\textbf{\textit{MLLM-based AD Diagnosis}}
Multimodal large language models (MLLMs) can handle heterogeneous inputs such as clinical text, images, and tabular data~\cite{haltaufderheide2024ethics,chen2023medblip,fang2024large}. In AD, earlier work mainly focused on single modalities, including structural MRI (sMRI)~\cite{frisoni2010clinical,jang2022m3t} or isolated clinical assessments~\cite{ohman2021current}. More recent methods integrate sMRI, cognitive scores, genetics, and biomarkers~\cite{venugopalan2021multimodal,chen2024alifuse,zhou2023transformer}, but typically output scalar predictions and remain opaque about how modalities contribute to decisions. With the emergence of MLLMs~\cite{openai2023gpt4,grattafiori2024llama}, several studies generate narratives from unimodal inputs (e.g., radiology reports from imaging~\cite{tian2023role,bai2024m3d}), yet they rarely combine diverse AD-related factors into a coherent, evidence-grounded diagnostic storyline. EMAD instead performs AD-specific multimodal reasoning: it jointly encodes 3D sMRI and structured clinical variables, applies bidirectional cross-modal fusion, and generates structured reports that explicitly integrate imaging and clinical evidence.

\noindent\textbf{\textit{Grounded and Anatomically Faithful Medical Reporting}}
Explainable medical AI has explored saliency maps, attention visualization, and prototype-based reasoning to interpret decisions~\cite{rajpurkar2022ai,chen2025holodx}. In neuroimaging, prior work often links classifier outputs to brain regions via post-hoc attribution or region-of-interest analyses~\cite{yao2023artificial}, while grounding in vision–language models has focused on phrase–region alignment in 2D images~\cite{chen2023medblip,wang2022medclip,wu2023medklip}. These approaches typically provide coarse, global explanations and do not connect individual report sentences to precise clinical fields and volumetric anatomy. Recent report-generation methods improve factuality by aligning text with structured EHR entries or radiology findings~\cite{gu2025radalign,miura2021improving,niu2024ehr,xie2019ehr}, but a hierarchical chain from sentence-level claims to clinical measurements and then to localized 3D MRI structures, especially for AD, remains underexplored. EMAD introduces Sentence--Evidence--Anatomy (SEA) Grounding, which first aligns each sentence to clinical evidence via multi-positive contrastive learning, then conditions a 3D segmentation decoder on the grounded evidence. To reduce annotation cost, GTX-Distill further transfers sentence$\rightarrow$evidence and evidence$\rightarrow$anatomy distributions from a teacher trained on a small grounded subset to a student operating on large-scale generated reports.

\noindent\textbf{\textit{Reinforcement Learning for Medical AI}}
Reinforcement learning (RL) is increasingly used to refine language models with preference- or rule-based rewards~\cite{schulman2017proximal,zhang2025survey,liu2025enhancing}. Group Relative Policy Optimization (GRPO)~\cite{shao2024deepseekmath} extends PPO by normalizing rewards within candidate groups and has shown benefits in text and vision–language settings~\cite{hu2025reinforce++,li2025optimizing}. In medicine, RL-style fine-tuning has been explored for improving factuality or enforcing simple templates in reports~\cite{zhou2021deep,lai2025med,pan2025medvlm,he2023blockchain}, but existing work is mostly unimodal and rarely encodes clinical guidelines as executable constraints. Reward functions often rely on heuristic similarity metrics that poorly capture clinical correctness, adherence to diagnostic criteria, or reasoning–conclusion consistency. In this work, we adapt GRPO to AD via an Executable-Rule RL-with-Verifiable-Rewards (RLVR) scheme: EMAD is optimized with rewards for structured format, NIA-AA–aligned diagnostic decisions, and natural-language-inference–based consistency between reasoning and final diagnosis, without relying on human preference labels.

%% file: sec/3_method.tex
\section{Methodology}

\subsection{Overall Framework}
We propose EMAD, an end-to-end vision-language framework for producing \textit{transparent, evidence-grounded, and anatomically faithful} medical image reports for Alzheimer's disease (AD) diagnosis. 
Given multimodal inputs $\mathcal{X}=\{x_v, x_t\}$, where $x_v \in \mathbb{R}^{D \times H \times W}$ denotes brain MRI scan and $x_t \in \mathbb{R}^{L}$ represents structured or semi-structured clinical variables (e.g., demographic, Genetic or Cognitive tests), the model consists of: 
(1) multimodal encoders; 
(2) projection and fusion layers; 
(3) a textual decoder for report generation; and 
(4) a hierarchical grounding head that enables explicit sentence-evidence-anatomy reasoning.
The overall framework is illustrated in Fig.~\ref{fig:overview}(a).

\noindent\textbf{\textit{Multimodal encoders}}
A visual encoder $E_v(\cdot)$ extracts patch-level visual embeddings $h_v$, while a textual encoder $E_t(\cdot)$ encodes textual features into $h_t$.

\noindent\textbf{\textit{Projection and Fusion}}
We first apply linear projections to map modality-specific features into the same embedding size
$h'_v,h'_t\in\mathbb{R}^{d}$.
To enable comprehensive interaction between neuroimaging and clinical modalities, we introduce an MFL comprising a Bidirectional Cross-Attention (BCA) mechanism. The projected features $h'_v$ and $h'_t$ are first processed by the BCA mechanism, where each modality alternately serves as Query and Key/Value to compute cross-attention:
$\mathbf{A}_{t \rightarrow v} = \text{Attn}(h'_t, h'_v, h'_v), \mathbf{A}_{v \rightarrow t} = \text{Attn}(h'_v, h'_t, h'_t).$
This bidirectional attention captures complex neuro-clinical dependencies, allowing visual features to inform clinical interpretation and vice versa. The attention outputs are combined with residual connections to preserve modality-specific information:$z_v = h'_v + \mathbf{A}_{v \rightarrow t}, z_t = h'_t + \mathbf{A}_{t \rightarrow v}$.

\noindent\textbf{\textit{Textual decoder}}
A causal language model $D(\cdot)$ generates a structured clinical report $\hat{\mathcal{R}}=\{\hat{s}_1,\ldots,\hat{s}_N\}$ conditioned on $z_v, z_t$.  
The final multimodal features $z_v$ and $z_t$ replace the placeholders \texttt{<sMRI>} and \texttt{<clinical>} in the input prompt templates. An example prompt for AD diagnosis is: 
\begin{quote}
\small
``Given the structural MRI \texttt{<sMRI>} and clinical profile \texttt{<clinical>}, what is the most probable diagnosis?''
\end{quote}

\subsection{SEA Grounding}
\label{sec:sea}
To ensure that every report claim is supported by both clinical and anatomical evidence, we introduce SEA Grounding (Fig.~\ref{fig:overview}(b)), a hierarchical alignment mechanism that performs: (1) sentence-to-evidence grounding, followed by (2) evidence-to-anatomy grounding.
During grounding training, SEA reuses the visual encoder $E_v$ and text encoder $E_t$, while only last linear layer remains trainable while others kept frozen. This design lets grounding supervision regularize only the task-specific top-layer representation toward evidence-centric features, while preserving the stability and efficiency of the pre-trained backbone.

\subsubsection{Sentence-to-Evidence Grounding}
For each generated sentence $\hat{s}_i$, we ground it directly to the \emph{input} clinical entries.
Let $\mathcal{E}=\{e_1,\dots,e_K\}$ be the set of clinical evidences parsed from inputs, where each $e_k=(t_k,a_k)$ contains a short textual descriptor $t_k$ (e.g., ``Hippocampal volume $4{,}724\,\mathrm{mm}^3$ '') and an attribute pointer $a_k$ that references the source field (e.g., an image-derived measurement).
We formulate sentence--evidence grounding as a many-to-many matching problem:
a sentence can be matched to multiple clinical evidences, and vice versa.
To handle such multi-positive cases, we adopt a multi-positive InfoNCE objective, applied symmetrically as in CLIP~\cite{radford2021learning}.
Let $\mathcal{P}_i$ and $\mathcal{N}_i$ denote the positive and negative sets for the $i$-th anchor (within a mini-batch).
Denote the sentence and evidence encoders by $E_t(\cdot)$, and define the scaled exponential similarity
$\kappa(s,e) \;=\; \exp\!\big(\mathrm{sim}(E_t(s), E_t(e))/\tau\big)$,
where $\mathrm{sim}(\cdot,\cdot)$ is cosine similarity and $\tau$ is a temperature.

The multi-positive InfoNCE loss is
\begin{equation}
\mathcal{L}_{\mathrm{SE}}
=\frac{1}{N}\sum_{i=1}^{N}\Big(\ell^{e\to s}_i+\ell^{s\to e}_i\Big),
\label{eq:ldg}
\end{equation}
with the evidence-to-sentence and sentence-to-evidence terms
\begin{equation}
\begin{aligned}
    \ell^{e\to s}_i
= \mathbb{E}_{\,j\in \mathcal{P}_i}\!
\left[
\log \frac{\kappa(e_i,s_j)}{\kappa(e_i,s_j)+\sum_{k\in \mathcal{N}_i}\kappa(e_i,s_k)}
\right], \\
\qquad
\ell^{s\to e}_i
= \mathbb{E}_{\,j\in \mathcal{P}_i}\!
\left[
\log \frac{\kappa(s_i,e_j)}{\kappa(s_i,e_j)+\sum_{k\in \mathcal{N}_i}\kappa(s_i,e_k)}
\right].
\end{aligned}
\label{eq:ldg_terms}
\end{equation}
This module does not require MRI inputs, only relies on sentence--clinical evidence pairs extracted from the inputs.

\subsubsection{Evidence-to-Anatomy Grounding}
If $e_{y_i}$ provides an anatomical pointer $a_{y_i}$, we localize it on the 3D MRI with an \emph{evidence-conditioned} 3D segmentation network~\cite{perera2024segformer3d} (Fig.~\ref{fig:overview}(b)).
Concretely, we start from a standard transformer-based 3D segmentation architecture and insert a lightweight \emph{cross-attention} block after the \emph{self-attention} in every decoder layer, allowing visual tokens to attend to the grounded evidence text.

Let $h_v\in\mathbb{R}^{S_v\times d}$ denote the sequence of visual tokens from the visual encoder, and let the grounded clinical evidence $e_{y_i}$ be encoded into a sequence of text tokens $t_i\in\mathbb{R}^{S_t\times d}$ via the same text encoder used in sentence--evidence grounding.
We use a $L$-layer transformer decoder.
For layer $\ell=1,\dots,L$, with residual connections and layer norms omitted for brevity, the update is:
\begin{align}
\mathbf{Y}^{(\ell)}_{\text{sa}} &= \mathrm{SelfAttn}\!\big(\mathbf{Y}^{(\ell-1)}\big), \\
\mathbf{Y}^{(\ell)}_{\text{ca}} &= \mathrm{CrossAttn}\!\big(\mathbf{Y}^{(\ell)}_{\text{sa}}\mathbf{W}^{(\ell)}_Q,\ 
t_i\mathbf{W}^{(\ell)}_K,\ 
t_i\mathbf{W}^{(\ell)}_V\big), \\
\mathbf{Y}^{(\ell)} &= \mathrm{FFN}\!\big(\mathbf{Y}^{(\ell)}_{\text{ca}}\big),
\end{align}
where $\mathbf{Y}^{(0)}=h_v$ and $\{\mathbf{W}^{(\ell)}_Q,\mathbf{W}^{(\ell)}_K,\mathbf{W}^{(\ell)}_V\}$ are learnable projections.
Intuitively, each decoder layer first aggregates long-range visual context (self-attention), then conditions on the grounded evidence by attending to $t_i$ (cross-attention).

The final decoder output $\mathbf{Y}^{(L)}$ is reshaped/upsampled and passed to a lightweight segmentation head to produce volumetric logits $\mathbf{P}_i\in\mathbb{R}^{H\times W\times D}$ and a probabilistic mask
\begin{equation}
\hat{\mathbf{M}}_i=\sigma\!\big(\mathrm{Head}(\mathbf{Y}^{(L)})\big),
\end{equation}
where $\sigma(\cdot)$ is the sigmoid.
Training uses the Dice + BCE objective adapted for masks, given a GT mask $\mathbf{M}_i$ for the structure,
\begin{equation}
\begin{aligned}
    \mathcal{L}_{\text{mask}} 
= &\lambda_{\mathrm{dice}}\!\left(1-\mathrm{Dice}(\hat{\mathbf{M}}_i,\mathbf{M}_i)\right) \\
+ &\lambda_{\mathrm{bce}}\mathrm{BCE}(\hat{\mathbf{M}}_i,\mathbf{M}_i).
\end{aligned}
\label{eq:sea_mask}
\end{equation}
This modification preserves the backbone design and adds only a small cross-attention block per decoder layer to inject text evidence, yielding explicit evidence-to-anatomy grounding.

\subsection{GTX-Distill: Grounding Transfer Distillation}

When all annotations are available, the model can be trained end-to-end as in Fig.~\ref{fig:overview}(a).
However, obtaining datasets with \emph{full} sentence--evidence and evidence--anatomy annotations is expensive.
To reduce dependence on dense labels, we propose GTX-Distill, a two-stage, label-efficient distillation strategy that aligns a student grounder to a teacher grounder (Fig.~\ref{fig:gtx}).

\noindent\textbf{\textit{Stage 1: Teacher grounder from grounding data}}
We train a \emph{teacher grounder} $G_{\mathrm{T}}$ using a small subset with grounding supervision.
The inputs are \emph{ground-truth} (GT) reports $\mathcal{R}=\{s_i\}$ paired with their clinical evidence links $\mathcal{E}_i\subseteq\mathcal{E}$ and anatomical masks $\{\mathbf{M}_i\}$.
$G_{\mathrm{T}}$ maps each sentence $s_i$ to a soft distribution over clinical evidences $q(e\,|\,s_i)$ and a soft anatomical mask $\mathbf{M}^T_i$.
The teacher is optimized with the supervised grounding losses introduced in Sec.~\ref{sec:sea}:
\begin{equation}
\mathcal{L}^{(T)}_{\text{GTX}}
=\mathcal{L}_{\text{SE}}^{\mathrm{T}}
+\lambda_{\mathrm{mask}}\mathcal{L}_{\text{mask}}^{\mathrm{T}}.
\label{eq:gtx_teacher}
\end{equation}

\noindent\textbf{\textit{Stage 2: Distillation to the student grounder}}
We \emph{freeze} $G_{\mathrm{T}}$ and train a \emph{student} grounder $G_{\theta}$ using \emph{generated} reports $\hat{\mathcal{R}}=\{\hat{s}_i\}$ from the textual decoder.
The teacher produces fixed evidence and anatomy distributions $q(\cdot\,|\,\hat{s}_i)$, while the student predicts $p_{\theta}(\cdot\,|\,\hat{s}_i)$.
We minimize a temperature-scaled KL divergence for evidence distributions:
\begin{equation}
\mathcal{L}^{\text{distill}}
=\tau^2 \sum_{i}\mathrm{KL}\!\left(q_{\tau}(\cdot\,|\,\hat{s}_i)\,\|\,p_{\theta,\tau}(\cdot\,|\,\hat{s}_i)\right).
\label{eq:gtx_kl}
\end{equation}

\noindent
The student-stage objective is therefore
\begin{equation}
\mathcal{L}^{(S)}_{\text{GTX}}
=\lambda_{\mathrm{KL}}\mathcal{L}^{\text{distill}}
\label{eq:gtx_student}
\end{equation}
optimized \emph{after} Eq.~\eqref{eq:gtx_teacher}, with $G_{\mathrm{T}}$ kept fixed and no gradient flowing into $q(\cdot)$.
We first minimize $\mathcal{L}^{(T)}_{\text{GTX}}$ w.r.t.\ teacher parameters; then, with the teacher frozen, we minimize $\mathcal{L}^{(S)}_{\text{GTX}}$ w.r.t.\ student parameters.
This sequential procedure transfers grounding knowledge from limited GT reports to model-generated reports, preserving sentence$\rightarrow$clinical and clinical$\rightarrow$anatomy faithfulness under limited supervision.

\subsection{Executable-Rule GRPO}
To enhance diagnostic faithfulness beyond SFT, we perform Reinforcement Fine-Tuning (RFT) with Group-Relative Policy Optimization (GRPO) under an RL-with-Verifiable-Rewards (RLVR) setup. 
The trainable components remain consistent with the SFT stage.
The total reward aggregates three executable components:
\begin{equation}
R \;=\; w_F\,R_F \;+\; w_{\text{NIA}}\,R_{\text{NIA-AA}} \;+\; w_C\,R_{\text{consistency}}.
\end{equation}

\noindent\textbf{\textit{Structured Format Reward}}
We enforce a minimal, machine-checkable schema: Reasoning, Diagnosis, Confidence. We set $R_F=1$ if all three tags exist and Confidence in [High, Medium, Low].

\noindent\textbf{\textit{NIA-AA Diagnostic Reward}}
To align with NIA-AA standards, we combine category alignment, biomarker consistency, and clinical feature coverage:
$R_{\text{NIA-AA}} \;=\; 0.4 \, R_{\text{cat}} \;+\; 0.3 \, R_{\text{bio}} \;+\; 0.3 \, R_{\text{feat}}.$
Category alignment $R_{\text{cat}}$ checks correct use of standardized labels (CN/MCI/Dementia) with exclusion of contradictions.
Biomarker consistency $R_{\text{bio}}$ scores coverage and status (normal/abnormal) for A$\beta$, tTau, pTau based on established thresholds.
Clinical feature coverage $R_{\text{feat}}$ rewards assessment across memory, executive function, visuospatial, and language domains with appropriate qualifiers.
Implementation details and exact scoring are in Appendix.

\noindent\textbf{\textit{Reasoning Consistency Reward}}
We quantify whether the final diagnosis is entailed by the preceding analysis via an NLI model~\cite{he2021debertav3}:
\begin{equation}
R_{\text{consistency}}
\;=\; \mathrm{NLI}\big(\text{Reasoning} \Rightarrow \text{Diagnosis}\big)
\end{equation}
corresponding to contradiction, neutral/weak entailment, and strong entailment, respectively. 
This prevents logically inconsistent outputs (e.g., normal biomarker descriptions followed by a “Dementia” diagnosis).

In conclusion, $R_F$ guarantees structural integrity, $R_{\text{NIA-AA}}$ enforces protocol-level clinical validity, and $R_{\text{consistency}}$ ensures that conclusions are justified by the model’s own reasoning. 
These verifiable rewards drive the model toward faithful, auditable AD reporting while maintaining stability via the KL regularizer to $\pi_{\text{ref}}$.

\subsection{Training Strategy}
We adopt a three-stage pipeline: Pre-Training (PT), Supervised Fine-Tuning (SFT), and Reinforcement Fine-Tuning (RFT)---to progressively endow EMAD with multimodal alignment, faithful grounding, and clinically verifiable reasoning for Alzheimer's disease.

\noindent\textbf{\textit{Stage 1: Pre-Training (PT)}}
To build modality-specific foundations and align imaging and clinical representations, we pre-train the visual encoder $E_v$ (sMRI) and the textual encoder $E_t$ on AD-relevant multimodal data. 

The optimization focuses exclusively on representation learning and alignment. We employ the image-text contrastive (ITC) loss~\cite{radford2021learning} to align image features $h_v$ and text features $h_t$ generated by the visual and text encoders. The ITC loss $\mathcal{L}_{\text{itc}}$ maximizes similarity for positive image-text pairs while suppressing negative pairs, implemented through normalized cross-entropy over all pairwise similarities.
We implement momentum encoders updated via exponential moving average (EMA) following BLIP~\cite{li2022blip}. 

Our restorative learning module is designed to enhance the global semantic understanding by incorporating fine-grained visual and textual information. 
That is, the feature extraction is augmented by a reconstruction learning branch, which includes an image decoder to reconstruct the original image from the representation and minimizes the pixel-level distance between the original image $x_v$ and the reconstructed image $x_v'$:
$\mathcal{L}_\text{res}^v =  \mathbb {E}_{x_v} \; \mathcal{D}_I(x_v, x_v')
$,
where $\mathcal{D}_v(x_v, x_v')$ presents the distance function that measures similarity between $x_v$ and $x_v'$, e.g., Mean Square Error (MSE), or L1 norm. We use MSE following the common setting~\cite{he2022masked}.
For the textual component, we apply a similar approach. 
A text decoder is trained to minimize the token-level distance between the original text $x_t$ and the reconstructed text $x'_t$:
$\mathcal{L}_\text{res}^t = \mathbb {E}_{x_t} \; \mathcal{D}_t(x_t, x_t')
$,
where $\mathcal{D}_t(x_t, x_t')$ is the distance function measuring text similarity, such as the commonly-used cross-entropy loss.
The overall pre-training objective combines both alignment and reconstruction losses:
$\mathcal{L}_{\text{PT}} = \mathcal{L}_{\text{itc}} + \lambda_{\text{res}} \left( \mathcal{L}_\text{res}^v + \mathcal{L}_\text{res}^t \right),$
where $\mathcal{L}_{\text{itc}}$ denotes image-text contrastive loss for feature alignment, $\mathcal{L}_\text{res}^v$ denotes image reconstruction loss (MSE), $\mathcal{L}_\text{res}^t$denotes text reconstruction loss (cross-entropy), $\lambda_{\text{res}}$ denotes weighting coefficient for reconstruction objectives.

\noindent\textbf{\textit{Stage 2: Supervised Fine-Tuning (SFT) + GTX-Distill}}
With all but the last layer of $E_v$ and $E_t$ frozen, we fine-tune their top layer together with the projection layers and the textual decoder (via LoRA) on AD diagnostic reports, while supervising grounding via SEA and distilling from a teacher (GTX-Distill). The text likelihood is
\begin{equation}
\mathcal{L}_{\text{txt}}
= -\,\mathbb{E}_{(\mathbf{T_Q},\,z_v,\,z_t,\,\mathbf{T_A})\sim\mathcal{D}}
\sum_{s=1}^{S}\log \pi_{\theta}\!\big(y_s \mid \mathbf{T_Q},\,z_v,\,z_t,\,y_{<s}\big),
\end{equation}
where $\pi_{\theta}$ is the decoder, $\mathbf{T_Q}$ the question/prompt, $\mathbf{T_A}$ the target answer tokens, and $(z_v,z_t)$ the fused features.

We attach SEA (Sec.~\ref{sec:sea}) for sentence$\rightarrow$clinical grounding and clinical$\rightarrow$anatomy localization, and apply GTX-Distill on generated sentences to match the teacher’s evidence/mask distributions:
\begin{equation}
\mathcal{L}_{\text{SFT}}
= \mathcal{L}_{\text{txt}}
+ \lambda_{\text{KL}}\,\mathcal{L}^{\text{distill}}
\end{equation}

\noindent\textbf{\textit{Stage 3: Reinforcement Fine-Tuning (RFT) with Executable-Rule GRPO}}
Finally, we apply GRPO with verifiable rewards to improve clinical faithfulness and protocol adherence (cf.\ Executable-Rule GRPO).
Given a query $\mathbf{Q}$ and sampled group $\{o_i\}_{i=1}^{G}$, we compute executable rewards and perform PPO-style group-relative updates with a KL anchor to the SFT reference:
\begin{equation}
\begin{aligned}
    \mathcal{L}_{\text{RFT}}
= -\,\frac{1}{G}\sum_{i=1}^{G}\min\!\Big(\rho_i A_i,\;\mathrm{clip}(\rho_i,1-\epsilon,1+\epsilon)A_i\Big) \\
+ \beta\,\mathrm{KL}\!\big(\pi_{\theta}(\cdot\mid\mathbf{Q})\,\|\,\pi_{\text{ref}}(\cdot\mid\mathbf{Q})\big)
\end{aligned}
\end{equation}
where $A_i$ is the group-normalized advantage from the executable reward, and $\rho_i$ is the importance ratio.
This stage optimizes clinically checkable behavior without additional human preference labels.

In conclusion, PT learns stable, shared embeddings; SFT injects task-specific reporting while enforcing sentence$\rightarrow$clinical$\rightarrow$anatomy faithfulness and distilling grounding; RFT further aligns outputs with executable clinical rules, yielding transparent and auditable AD reports.

%% file: sec/4_exp.tex
\section{Experiments}

\subsection{AD-MultiSense Dataset}

\begin{figure}[t]
    \centering
    \includegraphics[width=\linewidth]{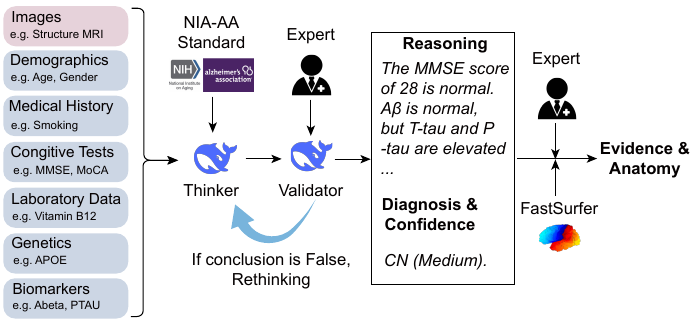}
    \caption{Pipeline for reasoning and grounding generation.}
    \label{fig:dataset}
\end{figure}

\textbf{Multimodal data collection.}
We construct AD-MultiSense from ADNI~\cite{petersen2010alzheimer} and AIBL~\cite{ellis2009australian} following established clinical logic~\cite{jack2018nia}. For each subject visit, we collect 3D sMRI and six categories of clinical data (demographics, history, cognition, laboratory tests, genetics, and biomarkers), yielding 10{,}378 multimodal samples from 2{,}619 subjects. Quantitative measurements are converted into standardized textual descriptors (z-normalized structural volumes mapped to atrophy levels, laboratory values filtered for significant abnormalities, biomarkers paired with reference-based interpretations), enabling language-based reasoning while preserving numerical information. Dataset statistics and full preprocessing details are provided in the Appendix.

\noindent\textbf{Reasoning and grounding generation.}
On top of these multimodal descriptors, we build diagnostic reports and grounding annotations (Fig.~\ref{fig:dataset}). A \textit{Thinker} model (DeepSeek-V3) is prompted as an AD specialist to produce structured outputs with \textit{Reasoning}, \textit{Diagnosis} (CN/MCI/Dementia), and \textit{Confidence}. A Validator module compares predictions with ground-truth labels and, using NIA-AA–based prompts, performs up to two refinement rounds; difficult cases receive the correct label and revised reasoning, yielding final pairs $\langle R^F, C^F \rangle$ for supervised training. In parallel, experts annotate clinical evidence links and 3D anatomical masks assisted by FastSurfer segmentation, providing sentence–evidence and evidence–anatomy supervision for SEA and GTX-Distill. Further details of the Thinker–Validator pipeline and grounding protocol are given in the Appendix.

\subsection{Experimental Setup}

All experiments are conducted on NVIDIA RTX 3090 GPUs. EMAD uses LLaMA 3.2-1B~\cite{grattafiori2024llama3herdmodels} as the text decoder with rank-8 LoRA adapters~\cite{hu2022lora}, a 3D Vision Transformer~\cite{dosovitskiy2020image} for sMRI, a Longformer encoder~\cite{beltagy2020longformer} for clinical text. Segformer3D~\cite{wu2022seqformer} used for SEA Grounding, we use 768-dimensional text embeddings with a contrastive temperature $\tau=0.07$, and a 4-layer 3D decoder. 
GTX-Distill employs a distillation temperature of 2.0 and KL weight $\lambda_{\mathrm{KL}}=1.0$. 
Executable-Rule GRPO is run with group size $G=4$, clipping parameter $\epsilon=0.2$, and KL coefficient $\beta=0.1$. 
Full optimization hyperparameters are provided in Appendix.

We access two binary tasks following clinical guidelines~\cite{mckhann2011diagnosis,dubois2007research,jack2018nia}: (1) distinguishing cognitively normal (CN) individuals from cognitively impaired (CI; including MCI and AD); and (2) distinguishing CN from MCI for early detection. In addition, we consider a three-way CN/MCI/AD classification task. The dataset is split \textit{subject-wise} into train/validation/test sets with ratios of 70\%, 10\%, and 20\%. All structural MRI scans are preprocessed by skull stripping~\cite{isensee2019automated} and intensity normalization~\cite{isensee2019automated}.

We evaluate EMAD from two perspectives: (i) report quality using BLEU, METEOR, ROUGE, and BERTScore; and (ii) diagnostic performance via accuracy (ACC), area under the ROC curve (AUC), sensitivity (SEN), and specificity (SPE).

\subsection{Overall Diagnostic and Reporting Performance}

\begin{table*}[htbp]
\scriptsize
\centering
\begin{tabular}{clcccccccc}
\toprule
& \textbf{Method} & \textbf{BLEU} & \textbf{METEOR} & \textbf{ROUGE} & \textbf{BERT} & \textbf{ACC} (\%) & \textbf{AUC} (\%) & \textbf{SEN} (\%) & \textbf{SPE} (\%) \\
\midrule
\multirow{11}{*}{CN vs. CI} 
& LLaVA-1.5-7B$^*$ & 0.0831 & 0.2417 & 0.2795 & 0.8012 & 74.23 & 70.58 & 62.14 & 82.36 \\
& LLaVA-Med$^*$    & 0.1024 & 0.2635 & 0.3042 & 0.8137 & 76.41 & 73.27 & 64.89 & 84.72 \\
& Med-PaLM-M$^*$   & 0.1189 & 0.2826 & 0.3314 & 0.8293 & 79.12 & 76.84 & 67.53 & 86.19 \\
& M3d-LaMed$^*$    & 0.1375 & 0.2982 & 0.3598 & 0.8341 & 82.37 & 79.65 & 70.94 & 87.56 \\
\cmidrule{2-10}
& LLaVA-1.5-7B & 0.2973 & 0.4764 & 0.5987 & 0.8485 & 86.42 & 83.19 & 80.37 & 88.54 \\
& LLaVA-Med    & 0.3186 & 0.4981 & 0.6179 & 0.8592 & 88.57 & 85.03 & 82.16 & 90.28 \\
& Med-PaLM-M   & 0.3394 & 0.5173 & 0.6371 & 0.8726 & 90.13 & 87.42 & 84.95 & 92.07 \\
& M3d-LaMed    & 0.3627 & 0.5419 & 0.6594 & 0.8748 & 91.28 & 89.16 & 86.72 & 93.14 \\
& EMAD (ours)  & \textbf{0.5422} & \textbf{0.6790} & \textbf{0.7781} & \textbf{0.9120} & \textbf{93.33} & \textbf{91.83} & \textbf{88.67} & \textbf{95.00} \\
\midrule
\multirow{11}{*}{CN vs. MCI} 
& LLaVA-1.5-7B$^*$ & 0.0715 & 0.2283 & 0.2594 & 0.7886 & 71.18 & 68.47 & 63.52 & 77.39 \\
& LLaVA-Med$^*$    & 0.0897 & 0.2472 & 0.2816 & 0.7991 & 73.42 & 70.59 & 66.84 & 79.21 \\
& Med-PaLM-M$^*$   & 0.1123 & 0.2698 & 0.3097 & 0.8184 & 76.35 & 73.48 & 68.92 & 82.17 \\
& M3d-LaMed$^*$    & 0.1294 & 0.2875 & 0.3391 & 0.8217 & 78.64 & 76.23 & 71.37 & 84.53 \\
\cmidrule{2-10}
& LLaVA-1.5-7B & 0.2826 & 0.4627 & 0.5789 & 0.8421 & 84.27 & 82.14 & 79.63 & 87.18 \\
& LLaVA-Med    & 0.3018 & 0.4815 & 0.6012 & 0.8534 & 86.39 & 84.27 & 81.45 & 89.32 \\
& Med-PaLM-M   & 0.3241 & 0.5036 & 0.6228 & 0.8649 & 88.21 & 86.45 & 83.72 & 91.08 \\
& M3d-LaMed    & 0.3437 & 0.5219 & 0.6413 & 0.8685 & 89.47 & 88.06 & 85.29 & 92.36 \\
& EMAD (ours)  & \textbf{0.5343} & \textbf{0.6421} & \textbf{0.7912} & \textbf{0.9130} & \textbf{92.82} & \textbf{90.09} & \textbf{88.60} & \textbf{93.50} \\
\bottomrule
\end{tabular}
\caption{Comparison of zero-shot ($^*$) and LoRA-finetuned baselines, and EMAD in terms of report quality and diagnostic performance for CN vs.\ CI and CN vs.\ MCI. }
\label{tab:final_results}
\end{table*}

\begin{figure*}[t]
    \centering
    \includegraphics[width=0.9\linewidth]{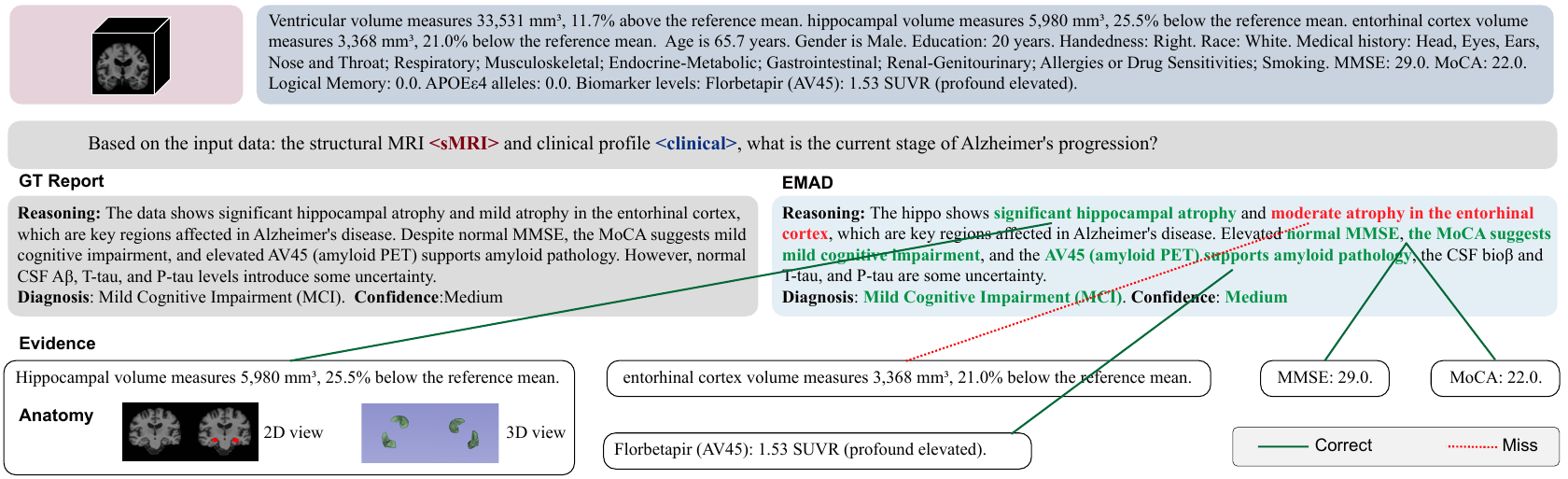} 
    \caption{Inference example of EMAD. The model integrates sMRI and clinical evidence to produce a grounded reasoning chain and diagnosis, with sentence-level claims linked to clinical fields and localized 3D anatomy.}
    \label{fig:example}
\end{figure*}


\begin{table}[htbp]
\centering
\scriptsize
\begin{tabular}{lccccc}
\toprule
\textbf{Method} & \textbf{ACC} & \textbf{Macro-F1} & \textbf{F1(CN)} & \textbf{F1(MCI)} & \textbf{F1(AD)} \\
\midrule
Longformer & 82.1 & 80.3 & 83.4 & 78.1 & 79.4 \\
Alifuse    & 84.7 & 82.5 & 86.9 & 80.2 & 80.5 \\
EMAD (ours) & \textbf{89.4} & \textbf{87.8} & \textbf{90.6} & \textbf{86.3} & \textbf{86.4} \\
\bottomrule
\end{tabular}
\caption{Three-way CN/MCI/AD classification performance (\%). }
\label{tab:three_class}
\end{table}

Table~\ref{tab:final_results} compares EMAD with four strong medical MLLMs on CN vs.\ CI and CN vs.\ MCI. EMAD achieves the best scores across all NLG and diagnostic metrics, outperforming M3D-LaMed by a large margin in both BERTScore and AUC. This indicates that explicit multimodal grounding improves not only narrative fidelity but also decision accuracy. 
To better reflect real-world AD staging, we also evaluate EMAD on three-way CN/MCI/AD diagnosis using the same subject-wise splits. We derive predicted labels from the generated reports and compare EMAD with the strongest multimodal baseline (Alifuse) and a text-only baseline.
Table~\ref{tab:three_class} summarizes the three-class results. Consistent with the binary tasks, EMAD achieves the best overall accuracy and macro-F1, with balanced performance across CN, MCI, and AD. In practice, we observe only a modest drop in ACC compared to the binary settings, suggesting that EMAD’s evidence-grounded reasoning scales well to more fine-grained staging.

We further compare EMAD against text-only and multimodal classification models~\cite{chen2024alifuse,liu2019roberta,beltagy2020longformer}. EMAD achieves higher ACC and AUC than all baselines. In addition, results for models trained on ADNI and evaluated on AIBL, demonstrating strong cross-cohort generalization, are also provided in the Appendix.

\subsection{Grounding Evaluation}

\begin{table}[htbp]
\centering
\scriptsize
\begin{tabular}{lccc}
\toprule
\textbf{Variant} & \textbf{R@1} & \textbf{R@3} & \textbf{MAP} \\
\midrule
No SEA (cosine only) & 0.42 & 0.61 & 0.53 \\
SEA w/o GTX          & 0.57 & 0.78 & 0.69 \\
SEA + GTX-Distill    & \textbf{0.65} & \textbf{0.84} & \textbf{0.76} \\
\bottomrule
\end{tabular}
\caption{Sentence-evidence grounding performance. }
\label{tab:sea_sentence}
\end{table}

\noindent\textbf{Sentence-evidence grounding.}
We first evaluate SEA Grounding in terms of how accurately it links generated sentences to clinical evidence fields. Given a sentence $s_i$ and the predicted evidence distribution $p(e \mid s_i)$, we treat annotated links as positives and compute retrieval metrics over all evidence candidates in the sample, reporting recall at $k$ (R@1, R@3) and mean average precision (MAP). We compare three variants: (i) a cosine-similarity baseline without SEA training, (ii) SEA without distillation, and (iii) SEA with GTX-Distill.
As shown in Table~\ref{tab:sea_sentence}, SEA significantly improves sentence--evidence alignment over the untrained cosine baseline, and GTX-Distill further boosts R@1/R@3 and MAP by transferring grounding behavior from a teacher trained on limited annotations.

\noindent\textbf{Evidence-anatomy grounding.}
We next assess evidence-to-anatomy grounding by comparing predicted 3D masks against expert-annotated masks (FastSurfer-assisted) for key AD-related regions (e.g., hippocampus, medial temporal lobe). We report Dice scores averaged over the test set.

\begin{table}[t]
\centering
\scriptsize
\begin{tabular}{lccc}
\toprule
\textbf{Model} & \textbf{Hippocampus} & \textbf{MTL} & \textbf{Overall} \\
\midrule
Image-only seg (no evidence) & 0.78 & 0.75 & 0.76 \\
Evidence-conditioned (EMAD)  & \textbf{0.84} & \textbf{0.81} & \textbf{0.82} \\
\bottomrule
\end{tabular}
\caption{Evidence--anatomy grounding Dice scores for key regions.}
\label{tab:sea_anatomy}
\end{table}

Conditioning the 3D decoder on clinical evidence improves Dice across all regions (Table~\ref{tab:sea_anatomy}), indicating that EMAD effectively ties textual findings (e.g., ``hippocampal atrophy'') to the corresponding anatomy in 3D sMRI.

\noindent\textbf{Label efficiency via GTX-Distill.}
Finally, we study label efficiency by varying the fraction of samples with full grounding annotations. We train (i) a teacher with supervision on the available subset and (ii) a student with or without GTX-Distill on the full set of generated reports. With only 25\% grounding labels, the GTX student retains about 95\% of the teacher’s sentence-evidence R@3, and with 50\% labels it essentially matches the fully supervised teacher shown in the Appendix.

\subsection{GRPO for Clinical Consistency}

To quantify the impact of Executable-Rule GRPO, we evaluate not only diagnostic accuracy but also several \emph{verifiable} properties of the generated reports: (i) structured format validity (presence of all \texttt{Reasoning/Diagnosis/Confidence} tags), (ii) NIA-AA diagnostic consistency (no contradiction between biomarker/status descriptions and the final label), and (iii) reasoning--diagnosis entailment measured by an NLI model.

\begin{table}[t]
\centering
\scriptsize
\begin{tabular}{lcccc}
\toprule
\textbf{Variant} & \textbf{ACC} & \textbf{Valid fmt} & \textbf{NIA-AA cons.} & \textbf{Entailment} \\
\midrule
EMAD w/o RFT      & 91.28 & 85.3 & 73.4 & 68.2 \\
+ Format reward   & 91.45 & 97.8 & 74.1 & 69.1 \\
+ Format + NIA-AA & 92.10 & 97.5 & 86.7 & 78.3 \\
EMAD & \textbf{92.82} & \textbf{99.1} & \textbf{90.8} & \textbf{87.6} \\
\bottomrule
\end{tabular}
\caption{Effect of Executable-Rule GRPO on CN vs.\ MCI. 
}
\label{tab:grpo}
\end{table}

As shown in Table~\ref{tab:grpo}, adding only the format reward mainly improves structural validity, while incorporating the NIA-AA diagnostic reward substantially increases guideline consistency without hurting accuracy. The full reward (including reasoning consistency) further improves both diagnostic ACC and entailment scores, demonstrating that Executable-Rule GRPO can steer EMAD toward clinically faithful and logically coherent reports without manual preference labels.

\subsection{Ablation Studies and Qualitative Analysis}

\noindent\textbf{Modality contributions.}
We further analyze the contribution of pre-training losses and modality fusion on CN vs.\ CI (Table~\ref{tab:ablation}). 
Using only sMRI or only clinical data leads to inferior performance; the best results are obtained when both modalities are present, underscoring the importance of multimodal integration. Finally, EMAD performs robustly under both IWG-2 and NIA-AA supervision, with NIA-AA providing slightly higher overall ACC/AUC and specificity.

\begin{table}[t]
\centering
\scriptsize
\begin{tabular}{lllll}
\toprule
    \textbf{Task} & \textbf{ACC} & \textbf{AUC} & \textbf{SEN} & \textbf{SPE} \\
\cmidrule(r){1-5}
    \textbf{(a) Feature Terms} \\
\cmidrule(r){1-5}
    Image & 71.24 & 54.76 & 95.33 & 12.31 \\
    Clinical & 88.83 & 82.69 & 96.91 & 67.42 \\
    Image + Clinical & \textbf{93.33} & \textbf{91.83} & \textbf{88.67} & \textbf{95.00}\\
\cmidrule(r){1-5}
    \textbf{(b) Guideline Terms} \\
\cmidrule(r){1-5}
    IWG-2 & 92.93 & 90.58 & \textbf{90.12} & 87.33 \\
    NIA-AA & \textbf{93.33} & \textbf{91.83} & 88.67 & \textbf{95.00}\\
\bottomrule
\end{tabular}
\caption{Ablation results (\%) on the CN vs.\ CI test set.}
\label{tab:ablation}
\end{table}

\noindent\textbf{Qualitative examples.}
Figure~\ref{fig:example} shows a representative EMAD prediction. The model jointly leverages sMRI and clinical data to justify its diagnosis, explicitly linking sentences to clinical fields (e.g., Hippocampus) and highlighting corresponding brain regions via 3D masks. We provide additional qualitative examples in the Appendix.

%% file: sec/5_conclusion.tex
\section{Conclusion}

We presented EMAD, an end-to-end vision–language framework for transparent, evidence-grounded AD diagnosis that jointly leverages 3D sMRI and rich clinical data on the AD-MultiSense dataset.
EMAD combines SEA Grounding (sentence-evidence-anatomy), GTX-Distill for label-efficient grounding transfer, and Executable-Rule GRPO with verifiable, NIA-AA–aligned rewards. Experiments show improved CN/MCI/AD staging and report quality, with explicit links from narratives to measurements and brain structures—moving toward reliable medical vision–language systems for early detection and monitoring.